\documentclass[sigconf,nonacm]{aamas}
\usepackage{balance} % for balancing columns on the final page
\usepackage{array} 
\usepackage[ruled,vlined,linesnumbered]{algorithm2e}

\setcopyright{ifaamas}
\acmConference[AAMAS '25]{Proc.\@ of the 24th International Conference
on Autonomous Agents and Multiagent Systems (AAMAS 2025)}{May 19 -- 23, 2025}
{Detroit, Michigan, USA}{A.~El~Fallah~Seghrouchni, Y.~Vorobeychik, S.~Das, A.~Nowe (eds.)}
\copyrightyear{2025}
\acmYear{2025}
\acmDOI{}
\acmPrice{}
\acmISBN{}

\acmSubmissionID{1385}

\title[AAMAS-2025 Formatting Instructions]{DPVS-Shapley:Faster and Universal Contribution Evaluation Component in Federated Learning}

\author{ Keting Yin}
\affiliation{
  \institution{Software College. Zhejiang University }
  \city{Hangzhou}
  \country{China}}
\email{yinkt@zju.edu.cn}

\author{Zonghao Guo}
\affiliation{
  \institution{Software College. Zhejiang University }
  \city{Hangzhou}
  \country{China}}
\email{guozonghao@zju.edu.cn}

\author{Zhenghan Qin}
\affiliation{
  \institution{Software College. Zhejiang University }
  \city{Hangzhou}
  \country{China}}
\email{leonine@zju.edu.cn}

\begin{abstract}
In the current era of artificial intelligence, federated learning has emerged as a novel approach to addressing data privacy concerns inherent in centralized learning paradigms.  This decentralized learning model not only mitigates the risk of data breaches but also enhances the system's scalability and robustness.  However, this approach introduces a new challenge: how to fairly and accurately assess the contribution of each participant.  Developing an effective contribution evaluation mechanism is crucial for federated learning.  Such a mechanism incentivizes participants to actively contribute their data and computational resources, thereby improving the overall performance of the federated learning system.  By allocating resources and rewards based on the size of the contributions, it ensures that each participant receives fair treatment, fostering sustained engagement.Currently, Shapley value-based methods are widely used to evaluate participants' contributions, with many researchers proposing modifications to adapt these methods to real-world scenarios. In this paper, we introduce a component called Dynamic Pruning Validation Set Shapley (DPVS-Shapley). This method accelerates the contribution assessment process by dynamically pruning the original dataset without compromising the evaluation's accuracy. Furthermore, this component can assign different weights to various samples, thereby allowing clients capable of distinguishing difficult examples to receive higher contribution scores.
\end{abstract}


\newcommand{\BibTeX}{\rm B\kern-.05em{\sc i\kern-.025em b}\kern-.08em\TeX}

\begin{document}

%%% The following commands remove the headers in your paper. For final 
%%% papers, these will be inserted during the pagination process.

\pagestyle{fancy}
\fancyhead{}

%%% The next command prints the information defined in the preamble.
\maketitle 
\section{Introduction}

Federated learning\cite{mcmahan2017communication} has emerged as a popular machine learning paradigm that allows for local data training, transmitting only model updates rather than raw data to a central server for aggregation. This approach ensures sensitive data remains within its domain, minimizing the risk of data breaches. While federated learning addresses issues such as data silos and privacy protection to some extent, a new challenge has arisen: how to fairly evaluate and compare the contributions of each client in a training session. Consequently, establishing a fair and effective incentive and reward allocation mechanism to encourage participants to provide higher quality data for circulation has become an urgent issue to resolve.
% 联邦学习作为作为一种热门的机器学习范式。联邦学习允许在本地进行数据训练，仅将模型更新而非原始数据传输到中央服务器进行聚合。这种方法确保敏感数据保持在其域内，从而最小化数据泄露的风险。虽然联邦学习在一定程度上解决了数据孤岛和隐私保护等问题，但如何公平评估、比较一次训练中每个客户端贡献的高低和多少已成为一个新的挑战。因此，建立一个公平有效的激励和奖励分配机制，以鼓励参与者提供更高质量的数据来参与数据流通，已成为一个亟待解决的问题。

The Shapley value\cite{shapley1953value} method has garnered significant attention from researchers due to its properties of group rationality, symmetry, null player, and additivity. However, it is also known for its high computational complexity and low efficiency. The evaluation process typically requires multiple rounds of validation on a validation set to accurately assess metrics such as accuracy. The original Shapley value approach, which required combining multi-party data and retraining the model, was extremely time-consuming. The introduction of MR\cite{song2019profit} technology and gradient aggregation-based methods eliminated the need for iterative model retraining, significantly improving the time required for measuring contributions. Nevertheless, these methods still face the issue of exponential growth in validation iterations due to combinatorial calculations. Existing work\cite{ghorbani2019data}\cite{liu2022gtg}\cite{jia2019towards} primarily focuses on reducing the number of validation rounds through sampling and truncation techniques, thereby shortening the overall validation time. Additionally, gradient aggregation-based Shapley value methods assign equal weight to each sample in the validation set, which is evidently unreasonable as the value of test samples often varies.
% 夏普利值法因其群体合理性、对称性、零元素和可加性等特性而备受研究人员的关注。 然而，它也以计算复杂度高和效率低而著称。 评估过程通常需要在验证集上进行多轮验证，以准确评估准确度等指标。 最初的夏普利值需要结合多方数据并重新训练模型，非常耗时。随着 MR 技术的引入，基于梯度聚合的方法无需对模型进行迭代式重新训练，从而大大改善了衡量贡献度所需的时间。但仍无法改变基于组合而产生的爆炸性验证次数问题，现有的工作主要是通过抽样和截断技术来减少验证轮数，从而缩短验证的总体时间。同时基于梯度聚合的夏普利值方法对于验证集中的每个样例赋予了相同的比重，这很显然是不合理的,测试样例之间的价值往往不是相等的。

To further enhance the efficiency of participant contribution assessment in Shapley value-based federated learning (FL), this paper proposes a method to accelerate the contribution evaluation process by reducing the time spent on each validation round, based on gradient aggregation techniques. We introduce the Dynamic Pruning Validation Set Shapley (DPVS-Shapley) method, which implements a dynamic pruning strategy on the validation set. By recording the results of the validation process, the validation set is divided into difficult and easy subsets. In each validation, extracting a certain proportion of cases from the easy subset and combining them with the difficult subset to form the validation set can achieve comparable accuracy to the full validation set while expediting the contribution assessment process. Furthermore, this method can assign different weights to the two types of samples, making the evaluation process more aligned with expectations and the assessment results more distinctive.
%为了进一步提高基于夏普利值的联邦学习（FL）中参与者贡献评估的效率，基于梯度聚合的方法，本文从另一个角度出发，提出了一种通过减少每轮验证所花费的时间来加速贡献评估过程的方法。我们引入了动态剪枝验证集夏普利（DPVS-Shapley）方法，该方法在验证集上实施动态剪枝策略。通过记录验证过程的结果，验证集被划分为难例集和易例集。在每次验证中，通过从易例集中提取一定比例的用例与难例集组成待验证集，可以获得与完整验证集相当的准确率，同时加快贡献评估的过程。此外，该方法还能为两种类型的样例分配不同的权重，使评估过程更符合预期，评估结果更有区分度。

Our main contributions are as follows: 1) We have discovered that by eliminating a subset of samples from the validation set, we can still achieve performance approximating that of the original validation set. 2) We conducted ablation experiments to demonstrate the effectiveness of our regression strategy and its variations. 3) Extensive experiments under various FL data distribution settings (i.i.d. and non-i.i.d. data distributions) show that DPVS can effectively improve the efficiency of contribution calculations without compromising the contribution assessment.

%我们的主要贡献如下：
%我们发现了通过删减一部分验证集中的用例依然可以拟合出近似原有验证集的效果。
%我们为基于梯度聚合和夏普利值方法提出了一种加速组件DPVS，该组件通过收集组合验证过程中的结果和中间参数。对验证集进行动态剪枝，从而减少每一轮的验证时间，最终达到降低计算成本的效果。此外，通过赋予相应的简单样例回归比例，该方法还可以为困难样例赋予更高的权重，从而使得实验结果更具有区分度。
%在各种 FL 数据分布设置（i.i.d. 和非 i.i.d. 数据分布）下进行的广泛实验表明，DPVS 可以在不影响贡献度评估的同时有效地提高了贡献计算的效率。
%%%%%%%%%%%%%%%%%%%%%%%%%%%%%%%%%%%%%%%%%%%%%%%%%%%%%%%%%%%%%%%%%%%%%%%%

\section{Related Work}
In recent years, the Shapley Value has also been widely applied in federated learning, primarily to assess the contribution of each participant to the training of the federated learning (FL) model.

The Shapley Value (SV) is a classical concept from game theory, introduced by Lloyd S. Shapley\cite{shapley1953value} in 1953, to measure the contribution of each player to the total payoff in cooperative games.However, calculating the Shapley Value involves considering various possible coalition combinations and permutations. As the number of participants increases, the computational burden grows significantly. For a game with n participants, the computational complexity can reach $ O(2^n) $. To address this issue, Castro et al.\cite{castro2009polynomial} proposed a Monte Carlo (MC) sampling method to approximate the Shapley Value with fewer utility evaluations, achieving favorable results. Following this, Maleki et al.\cite{maleki2013bounding} explored the complexity of approximating the Shapley Value (SV) using independent Monte Carlo methods, providing theoretical insights and proofs.Van Campen et al.\cite{van2018new}introduced the SMC-Shapley method, which first applied the Shapley Value framework to federated learning.  They proposed a structured Monte Carlo sampling estimation method to effectively approximate the Shapley Value.  Given the irregular utility values of different coalitions, their method approximates the Shapley Value by swapping participants' positions within coalition sequences.Ghorbani et al. \cite{ghorbani2019data} proposed the TMC-Shapley method, which calculates the Shapley Value for each data point by sequentially adding data points.  At each addition, the algorithm computes the performance difference between the current and previous subsets.  If the difference is below a pre-set performance tolerance threshold, the algorithm halts, thus achieving early termination and further reducing the computational load.Jia et al. \cite{jia2019towards} proposed the GTB-Shapley method.  This method employs the concept of "group testing," where participants are divided into multiple groups for testing, thereby reducing the number of utility function evaluations.
% 夏普利值（SV）是一个源自博弈论的经典概念，由劳埃德·S·夏普利于1953年提出，用于衡量每个参与者对合作游戏总收益的贡献。近年来，夏普利值也被广泛应用于联邦学习，主要用于评估每个参与者对联邦学习（FL）模型训练的贡献。然而，计算夏普利值需要考虑各种可能的联盟组合和排列。随着参与者数量的增加，计算负担显著增加。对于具有 n 个参与者的联邦计算，计算复杂度可以达到O(2^n)。当n从5增加到10时，就意味着复杂度从32增长到了1024。这代表利用夏普利值计算单轮贡献时需要在一个验证集中验证的次数从32次增长到了1024次。为了解决这个问题，Castro 等人提出了一种蒙特卡洛（MC）采样方法，通过减少效用评估次数来近似夏普利值，取得了良好的效果。随后，Maleki 等人探讨了使用独立的蒙特卡洛方法近似夏普利值的复杂性，并提供了理论见解和证明。Van Campen 等人引入了 SMC-Shapley 方法，这是首次将夏普利值框架应用于联邦学习。他们提出了一种结构化的蒙特卡洛采样估计方法，有效地近似夏普利值。鉴于不同联盟的效用值不规则，他们的方法通过在联盟序列中交换参与者的位置来近似夏普利值。Ghorbani 等人提出了 TMC-Shapley 方法，该方法通过顺序添加数据点来计算每个数据点的夏普利值。在每次添加时，算法计算当前子集和前一个子集之间的性能差异。如果差异低于预设的性能容忍阈值，算法将停止，从而实现提前终止，进一步减少计算负担。Jia 等人提出了 GTB-Shapley 方法。该方法采用“组测试”的概念，将参与者划分为多个组进行测试，从而减少效用函数评估的次数。

Sampling-based methods effectively reduce the cost of the validation process. However, earlier approaches often require retraining the model, leading to significant time spent on repeated model training. Song et al. \cite{song2019profit} proposed reconstructing the model based on gradients provided by participants, thereby avoiding model retraining and reducing the exponential cost.Jia et al.\cite{jia2019efficient} introduced KNN-Shapley from a data pruning perspective, which employs K-nearest neighbors in federated tasks to perform data sample pruning for contribution assessment. However, this method exhibits considerable deviation compared to the true Shapley Value.Ghorbani et al. \cite{ghorbani2020distributional} explored scenarios with a large number of participants, where their method samples only a few participants to optimize the efficiency of contribution evaluation. The contributions of unsampled participants are then assessed using regression fitting.Liu et al. \cite{liu2022gtg} accelerated convergence and reduced computational costs by using guided permutation sampling combined with both inter-round and intra-round truncation.

% 基于采样的方法有效降低了验证过程的成本。然而，早期的方法通常需要重新训练模型，导致重复模型训练上花费了大量时间。Song 等人提出了基于参与者提供的梯度重构模型的方法，从而避免了模型重新训练并降低了指数级的成本。Jia 等人从数据剪枝的角度引入了 KNN-Shapley，该方法在联邦任务中使用 K 最近邻进行数据样例剪枝以评估贡献。然而，该方法与真实的沙普利值相比存在相当大的偏差。Ghorbani 等人探讨了参与者数量较多的场景，他们的方法仅对少数参与者进行采样，以优化贡献评估的效率。未采样参与者的贡献则通过回归拟合进行评估。Liu 等人通过结合跨轮和轮内截断的引导排列采样，加速了收敛并降低了计算成本。

Furthermore, some researchers have utilized utility function approximation methods to calculate contributions. Wang et al.\cite{wang2020principled} proposed FedSV, which approximates the utility function by randomly selecting a group of clients, setting the contributions of other clients to zero in the current round. Fan et al.\cite{fan2022improving} introduced a novel contribution evaluation algorithm, ComFedSV, employing a low-rank matrix completion model to assess Shapley value-based contributions. This method records the improvement in the loss function between updated model parameters and those from the previous round. However, it is only applicable to horizontal federated learning. To address this limitation, Fan et al. subsequently proposed VerFedSV\cite{fan2022fair} for vertical federated learning. This approach is suitable for both synchronous and asynchronous settings, evaluating data contributions while measuring communication and computational performance. Nevertheless, calculating Shapley values still requires considerable time. Other researchers have combined distributed computing, edge computing, homomorphic encryption, and other technologies with Shapley values. Ma et al.\cite{ma2021transparent} proposed a blockchain-based federated learning framework and a protocol for transparently evaluating each participant's contribution. Liu et al.\cite{liu2020fedcoin} introduced a peer-to-peer payment system called FedCoin, utilizing a "Proof of Shapley" (PoSap) consensus protocol to compute the Shapley value for each data owner, as opposed to traditional proof-of-work methods. Dong et al.\cite{dong2023affordable} presented an efficient Shapley value estimation method, leveraging the advantages of edge computing to achieve an affordable federated edge learning framework. Zheng et al.\cite{zheng2022secure} proposed HESV, a homomorphic encryption-based single-server solution, and SecSV, a dual-server solution, to address security issues in federated learning systems.

% 此外，有的研究者利用基于效用函数逼近的方法来计算贡献度。Wang等人提出了FedSV，该方法通过随机选择一组客户端来近似效用函数，而使得其他客户端的贡献在本轮中被设为。Fan等人提出了一种新的贡献评估算法 ComFedSV，使用低秩矩阵补全模型来评估基于沙普利值的贡献，记录更新模型参数与上一轮模型参数对损失函数的改善，但该方法只适用于水平联邦学习。为此Fan等人又为垂直联邦学习提出了 VerFedSV，该方法适用于同步和异步设置，评估数据贡献的同时衡量通信和计算性能，但计算夏普利值依旧需要较长的时间。还有的研究者将分布式、边缘计算、同态加密等技术和夏普利值相结合。Ma等人提出了一个基于区块链的联邦学习框架和一个透明地评估每个参与者贡献的协议。Liu等人提出了一种称为 FedCoin 的点对点支付系统，使用“夏普利值证明”（Proof of Shapley, PoSap）共识协议计算每个数据所有者的夏普利值，而不是传统的工作量证明。Dong等人提出了一种高效的夏普利值估计方法，结合了边缘计算的优势，旨在实现可负担的联邦边缘学习框架。Zheng等人提出基于同态加密的单服务器解决方案 HESV 和双服务器解决方案 SecSV，以解决联邦学习系统中的安全问题。

\section{Preliminaries}
\subsection{Shapley Value In Federated Learning}
The Shapley Value (SV) is a classical concept from game theory, introduced by Lloyd S. Shapley\cite{shapley1953value} in 1953, to measure the contribution of each player to the total payoff in cooperative games. 

In federated learning, the Shapley Value is defined as follows: Consider $n$ participants with datasets $ D_1,D_2,...,D_{n-1},D_n $,a model $M$,and a standard test set $T$.Let $D_S$ denote a combined dataset where $ S \subseteq N $.The model $M$ trained on the dataset $D_S$ is denoted as $M_S$.The performance of the model on the standard test set is denoted by $U(M,T)$,abbreviated as $ U(M) $.The Shapley Value $ \phi(D_N,T,D_i) $,abb\-reviated as $ \phi_i $, is used to compute the contribution of each FL participant $i$, and is defined by the formula:
$$
\phi_i = C \sum_{S\subseteq N \setminus \{i\}} \frac{U(M_{S\bigcup\{i\}})-U(M_S)}{\dbinom{n-1}{|S|}}
$$

Where C is a constant. Obviously, calculating Shapley requires a lot of retraining of the model, which creates a huge overhead, which is unacceptable for individual clients.

\subsection{Optimization Strategies}
Before computing Shapley values based on model aggregation, obtaining model performance metrics often requires retraining the models. This can lead to significant time spent on training models based on different data combinations, which is clearly unacceptable for participants, as they do not gain additional rewards from the evaluation process. The proposals of MR\cite{song2019profit} address this issue by accumulating the contribution of each new round of evaluation and using the model, which aggregates client gradients, as the new base model, thus solving the problem of model retraining. This shift moves the primary time-consuming part of contribution evaluation from model training to model evaluation.

Currently, most methods rely on sampling\cite{wang2019measure} and early stopping to achieve the convergence of the convergence function and ultimately fit the true Shapley values. These methods primarily aim to accelerate by reducing the number of validations on the validation set. From another perspective, acceleration can also be achieved by attempting to reduce the time of each individual validation, which aligns with the concept of data pruning. By reducing the amount of data that needs to be validated, corresponding speed improvements can be achieved. Based on local correlation characteristics, federated learning uses K-nearest neighbor task models to achieve data sample pruning for contribution evaluation\cite{jia2019efficient}.
\subsection{Model Stability}

Support Vector Machine (SVM)\cite{jakkula2006tutorial}\cite{yu2012svm} is a machine learning algorithm used for classification and regression analysis, with its primary application in classification problems. The core concept of SVM is to identify an optimal decision boundary (hyperplane) that separates data points into different categories while maximizing the margin between classes. As the learning process progresses, changes in the classification boundary become increasingly subtle, with most samples consistently remaining on one side of the hyperplane. Points closer to the decision boundary have a higher likelihood of being classified to the opposite side in the next round of model training. During the validation process, we tend to focus more on the points that are misclassified in the current round and those that are likely to be misclassified in the next, while paying relatively less attention to the points that are stably classified on either side.

% 支持向量机（SVM）是一种用于分类和回归分析的机器学习算法，最常用于分类问题。SVM的核心思想是找到一个最佳决策边界（超平面），将数据点划分为不同的类别，同时最大化类别之间的间隔。在学习过程中，分类边界的变化变得越来越小，大多数样例始终保持在超平面的任一侧。越靠近决策边界的点，在下一轮模型训练中被划分到另一侧的可能性越大。在验证的过程中，我们往往更加关注与这一轮被分错的点以及将要分错的点，而相对不那么关心那些稳定被划分在两侧的点。

The universal approximation capability of neural networks was demonstrated by Cybenko et al.\cite{cybenko1989approximation}, showing that single-hidden-layer neural networks can approximate any continuous function. Hornik et al.\cite{hornik1989multilayer} further explored the universal approximation capabilities of multi-layer feedforward neural networks, explaining why neural networks can learn complex decision boundaries. Neural networks leverage their hierarchical structure and non-linear activation functions to learn highly complex decision boundaries, enabling sophisticated mappings from input to output.

Although the decision boundaries of neural networks are often highly non-linear, sample classification tends to stabilize as the model is trained. We have adapted our SVM-inspired approach to neural networks, considering samples that are consistently classified correctly and far from the decision boundary as candidates for pruning. This approach allows us to progressively reduce the number of samples being tested.

\section{The Proposed Approach}

\subsection{Main Process}
\begin{figure*}[h]
    \centering
    \includegraphics[width=\textwidth]{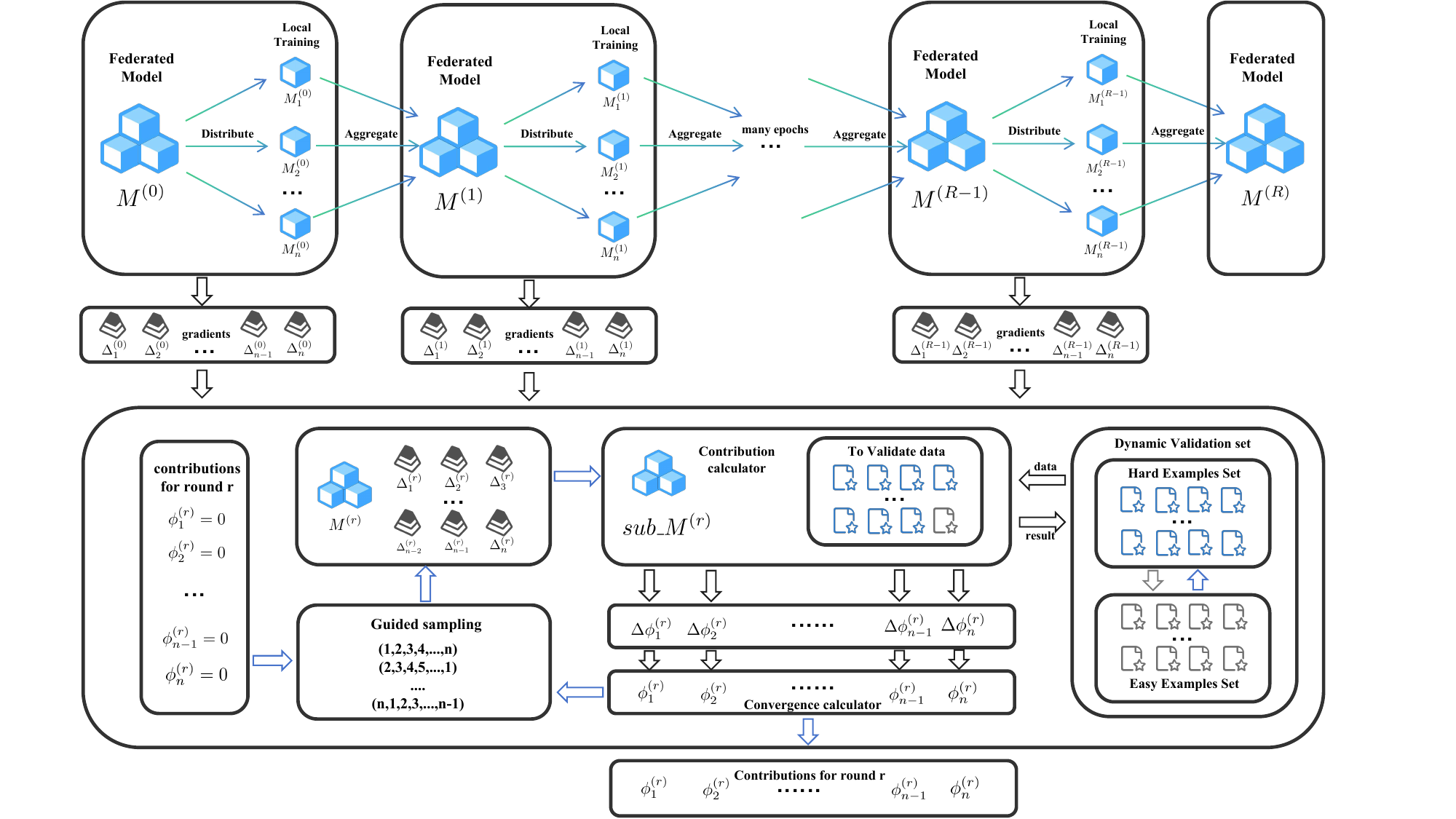}
    \caption{DVPS-Shapely}
    \label{fig:DVPS-Shapely}
\end{figure*}

Based on the existing problems and related explorations, we propose a method for validation set pruning aimed at accelerating the single-round validation process by reducing the number of validation samples, thereby speeding up the entire contribution assessment process. Figure \ref{fig:DVPS-Shapely} illustrates the confidence regression-based flowchart. In federated learning, the server distributes the model, and clients upload gradients to the server after local training. Contribution assessment methods based on gradient combination using Shapley value require repeatedly combining gradients and evaluating on a validation set. The final contribution proportions are then calculated based on marginal benefits derived from all validation results.
% 基于现有问题和相关探索，我们提出了一种针对验证集剪枝的方法，旨在通过减少验证集的数量来加速单轮验证过程，从而达到加速整个贡献度评估的过程，如图 \ref{fig:DVPS-Shapely} 所示展示的基于置信度回归的流程图。在联邦学习中，服务器会下发模型，客户端在本地训练后会将梯度上传到服务端。在基于梯度组合的夏普利值贡献度计算方法中，需要反复组合梯度之后再在验证集上进行验证，最终基于所有的验证结果计算边际效益得到最后的贡献度占比。

Our method comprises two phases.The first phase is the experience accumulation phase, during which no pruning is performed on the validation set. This is because the model is still relatively unstable at this stage and unable to effectively distinguish between simple and difficult examples. By evaluating the model on the complete validation set, we accumulate multiple rounds of sample decision results and decision confidence scores. Once the accumulated quantity meets the requirements, we can classify the samples into simple and difficult examples, allowing us to enter the dynamic pruning phase.
% 我们的方法包括两个阶段。第一阶段是经验积累阶段，在此期间不对验证集进行剪枝，因为此时模型还相对不稳定，无法有效的区分简单样例以及困难样例。通过在完整的验证集上评估模型，我们积累了多轮样例决策结果和决策置信度得分。一旦积累的数量满足要求，我们就可以按划分出简单样例以及困难样例，我们就可以进入动态剪枝阶段。

The second phase is the dynamic pruning phase. In this stage, we first categorize the samples in the validation set into difficult and simple examples based on previous sample judgment results. According to our extraction strategy, we select a corresponding proportion of simple examples and combine them with all difficult examples to obtain the pruned test set. If a confidence-based extraction strategy is chosen, the probability of selecting samples with lower confidence increases during the extraction process. We believe that these samples are closer to the decision boundary, and their judgment results are more likely to change in the model's next training process.The pruning algorithm is shown in Algorithm \ref{algorithm:1}.
% 第二阶段是动态剪枝阶段。在这一阶段，我们首先根据之前的样例判断结果，将验证集中的样例分为困难样例和简单样例。根据我们的抽取策略，我们会选择相应比例的简单样例，并将它们与所有困难样例合并，得到剪枝后的测试集。如果选择基于置信度的抽取策略，那么在抽取过程中，选择置信度较低样例的概率就会增加。我们认为，这些样例更接近决策边界，其判断结果更有可能在模型的下一个训练过程中发生变化，剪枝算法如算法2所示。

We run the aggregated model on this dynamically pruned validation set to obtain the model's accuracy on the pruned test set and collect intermediate results. Since our accuracy does not include the pruned samples, we need to adjust this accuracy accordingly. The final accuracy formula \ref{eq:acc} is shown as follows.
% 在这个动态剪枝验证集上运行聚合模型以获取模型在剪枝后的测试集的准确率，并收集中间结果。由与我们的准确率不包含被剪枝部分的样例，所以我们需要对该正确率进行相应的调整，最终准确率的如下公式所示。

\begin{equation}
\begin{split}
  Acc_{final} = {(100*\frac{N_{prune}}{N_{sum}}+ Acc_{dynamic}*\frac{N_{dynamic}}{N_{sum}})}
  \label{eq:acc} % 给公式一个标签
\end{split}
\end{equation}

As the results collected during the validation process only include unpruned test samples, we need to supplement the validation results and confidence scores of the pruned data using historical records. For the judgment results of pruned samples, we assume all are correct by default. However, for confidence scores, we apply a certain confidence decay, because as the model training rounds accumulate, doubts arise about previous classification results. Reducing confidence increases the probability of the sample being selected in the next extraction, thereby enabling timely and effective regression. The regression algorithm is shown in Algorithm \ref{algorithm:2}.
% 由于我们在验证过程中收集到的结果只包括未剪枝的测试样例，因此我们需要用历史记录来补充被剪枝的数据的验证结果以及置信度。对于剪枝样例的判断结果，我们默认全部正确。然而，对于置信度分数，我们会进行一定的置信度衰减，因为随着模型训练轮次的叠加，对于之前的分类结果的判断会产生怀疑，降低置信度有利于增加该样本下一次抽取时被选中的概率，从而可以及时有效的进行回归，回归算法如算法2所示。

\begin{algorithm}
\caption{Dynamic pruning algorithm}
\label{algorithm:1}
\KwIn{numEpoch, startDynamicEpoch, sampleRate, n, m, judgeResultMatrix, judgeConfidenceMatrix, allValidationCases, randomType}
\KwOut{needValidationCases}
\eIf{nowEpoch $<$ startDynamicEpoch}{
    \Return allValidationCases\;
}{
    \ForEach{valCase $\in$ allValidationCases}{
        \eIf{valCase latest n epoch is all correct in judgeResultMatrix}{
            add valCase to easySet\;
        }{
            add valCase to hardSet\;
        }
    }
    \uIf{randomType $==$ None}{
        \Return easySet\;
    }
    \uElseIf{randomType $==$ Random}{
        selectSet $\leftarrow$ randomly select n percent of the samples from the easySet\;
        \Return selectSet $\cup$ hardSet\;
    }
    \ElseIf{randomType $==$ WeightRandom}{
        weights $\leftarrow$ []\;
        \ForEach{easyCase $\in$ easySet}{
            aveConfidence $\leftarrow$ average the confidence level of the last m rounds of this easyCase\;
            add aveConfidence to weights\;
        }
        selectSet $\leftarrow$ randomly select n percent of the samples from the easySet based on weights\;
        \Return selectSet $\cup$ hardSet\;
    }
}
\end{algorithm}

\begin{algorithm}
\caption{Information updating algorithm}
\label{algorithm:2}
\KwIn{numEpoch, startDynamicEpoch, judgeResultRecord, judgeResultMatrix, confidenceResultRecord, confidenceResultMatrix, beta}
\KwOut{judgeResultMatrix, confidenceResultMatrix}

\If{nowEpoch $<$ startDynamicEpoch}{
    add judgeResultRecord to judgeResultMatrix\;
    add confidenceResultRecord to confidenceResultMatrix\;
    \Return\;
}
allJudgetRecord $\leftarrow$ []\;
allConfRecord $\leftarrow$ confidenceResultMatrix[-1]\;
\For{i $\leftarrow$ 0 \KwTo testCaseNum - 1}{
    \eIf{judgeResultRecord contains the result of the i-th sample}{
        allJudgetRecord[i] $\leftarrow$ judgeResultRecord[i]\;
    }{
        allJudgetRecord[i] $\leftarrow$ 1\;
    }
    \eIf{confidenceResultRecord contains the confidence of the i-th sample}{
        allConfRecord[i] $\leftarrow$ confidenceResultRecord[i]\;
    }{
        allConfRecord[i] $\leftarrow$ allConfRecord[i] $\times$ beta\;
    }
}
add allJudgetRecord to judgeResultMatrix\;
add allConfRecord to confidenceResultMatrix\;
\Return\;
\end{algorithm}

\subsection{Why and how to prune simple sample}
To further reduce the time spent on contribution evaluation, we propose the concept of Dynamic Pruned Validation Set (DPVS) from the perspective of decreasing the validation time per round. The question of which examples to include in the pruned dataset arises. We posit that samples that do not affect accuracy should be eliminated, specifically those consistently classified correctly during the model's decision process. Such samples have negligible impact on accuracy, as we can largely anticipate their correct classification in the subsequent validation round. 
To further reduce the time consumed by contribution assessment, we propose the concept of Dynamic Pruning Validation Set (DPVS) from the perspective of decreasing validation time per round. The question of which examples in the pruned dataset to retain becomes a new challenge. We argue that samples that do not affect accuracy should be removed, specifically those consistently classified correctly during the model's discrimination process. Such examples have minimal impact on accuracy, as we can largely anticipate their correct classification in subsequent validation rounds. We define these as simple samples.
To this end, we collect the classification results for each round, $R_{sample}=[0,1,1,....,1,0,1]$, which indicates whether each sample was correctly classified. After n rounds, we obtain a two-dimensional matrix containing the validation results. This matrix allows us to categorize the validation set samples into simple and difficult samples. We define simple samples as follows: if a sample is correctly classified in all of the most recent n rounds, it is considered an simple sample; otherwise, it is deemed a difficult sample. In our subsequent experiment, "Simple sample scale in different dataset", we found that the proportion of simple samples varies from approximately 30\%+ to 90\%+ across different datasets. This finding suggests that our categorization is meaningful and can potentially achieve corresponding proportions of validation acceleration. By focusing solely on predicting difficult samples, we can achieve faster validation speeds. However, considering that models undergo continuous changes during the training process, which may lead to previously simple samples becoming misclassified, it is necessary to consider how to timely and effectively reintegrate these samples into the validation set.

% 为了进一步减少贡献度评估的所耗费的时间，我们从减少单轮次的验证时间的角度出发，提出了动态剪枝数据集（DPVS）的概念。剪枝数据集中的哪些样例称为了新的问题。我们认为应该减去那些不影响准确率的样例，就是那些在模型判别过程中一直被判别为正确，这样的样例对于准确率的影响几乎是无，因为我们可以在很大程度上预料在下一轮的验证中该样例依旧被判定为正确,我们将这一类的样本定义为简单样例。为此我们收集每轮中的判定结果R_{sample}=[0,1,1,....,1,0,1]，包含每个样本是否被判断正确的标记。收集n轮之后就可以得到一个包含n轮验证结果的二维矩阵。通过这个二维矩阵我们可以将验证集中的样例划分为简单样例和困难样例，我们对于简单样例的定义如下：如果样例在最近的n轮次中均被正确判断，则该样例被视为简单样例；否则，样例被视为难样例。我们在之后的实验“不同数据集中简单样例的规模”实验中发现，不同数据集中简单样例存在的比例从30%左右到90%左右不等，这意味着我们的切分是有意义的，可以获得对应比例的验证加速。我们只需要关注于难例样例的预测即可获得更快的验证速度，但考虑到模型在训练过程中会不断产生变化，可能导致原有的简单样例无法再被正确识别，这就需要考虑到如何对简单样例进行及时和有效的回归。

\subsection{Strategies for Updating Dynamically Pruned Validation Sets}
To address this issue, we propose two methods: random regression and confidence-based regression. Random regression involves randomly selecting a certain proportion of samples from the simple samples to combine with the difficult examples, forming the validation set. Each simple sample has an equal probability of being selected, while the remaining simple samples constitute the pruned portion. The confidence-based regression method requires collecting not only the classification results but also the confidence levels associated with these results. We apply a function transformation to the confidence levels, ensuring that simple samples with higher confidence have a lower probability of being selected during the sampling process.
%为此我们提出了随机回归和基于置信度的回归两种方式，随机回归是指从简单的样例中随机抽取一定比例的样例连通困难样例组成待验证的测试集，每个简单样例被抽取到的概率相等，而剩下的那部分简单样例就是被剪枝的部分。基于置信度的回归方式需要在收集判定结果的同时收集对于该结果的置信度，同时我们会对置信度进行相应的函数变化使得置信度高的简单样例在抽取时被抽取到的概率会更低。

To address different objectives, we propose two update timing strategies. The EE strategy aims to update the validation set after each validation round, ensuring that the accuracy obtained from this set more closely approximates the true accuracy. However, in our subsequent experiment "Effect of aggregated gradient number on the accuracy of pruning data set", we discovered that when the number of combined gradients is low, the resulting aggregated model becomes too similar to a single client's model. This leads to significant discrepancies in accuracy. Consequently, we found it necessary to increase the sampling ratio when dealing with a smaller number of combined gradients.To assign greater importance to difficult samples and further accelerate contribution measurement, we propose the ET strategy. This strategy updates the validation set only before contribution calculation, ensuring that all gradient-combined models are validated on the same set. By adopting this strategy, we no longer focus on fitting the contribution of the original Multi-Round (MR) approach, allowing us to use a smaller regression ratio. The purpose of this strategy is to widen the gap between different parties' contributions, enabling clients capable of identifying difficult samples to have a larger contribution share. This approach increases the disparity between parties, enhancing differentiation and facilitating effective contribution ranking.

%针对不同的目标，我们提出了两种更新时机策略。EE策略旨在每次验证完就更新新的验证集使得从该验证集上得到的准确率会更接近真实的准确率，但在后续的实验“不同梯度组合数对于准确率的影响”中我们发现，当组合的梯度数较少时，使得我们组合后端模型过于接近某一个客户端的模型，从而导致准确率的差距过大，为此我们不得不加大组合数较少时的采样比例。同时为了赋予困难样例更大的比例以及进一步加速贡献度的衡量，我们提出了ET的更新策略，该策略只在贡献度计算之前更新验证集，使得所有梯度组合的模型均在同一验证集上进行验证。采用该策略不在执着于拟合原有MR的贡献度，这使得我可以使用较小的回归比例。该策略的目的在于拉大各方贡献度的差距，使得能够识别困难样例的客户端拥有更大的贡献占比，拉大各方差距，提高区分度，便于进行有效的贡献度排名。

\section{Experimental Evaluation}
\subsection{Settings}
The hardware and software environments for the experiment are as follows:
\begin{itemize}
    \item Operating system: Ubuntu 16
    \item Development language: Python 3.9
    \item CPU: Intel Xeon Gold 5118
    \item GPU: NVIDIA RTX A4000, 16GB of video memory.
    \item Memory: 96GB
    \item Hard disk: 3.6TB capacity
\end{itemize}

This experiment employs the FedAvg (Federated Averaging) strategy for federated aggregation. The dataset used in the experiments is CIFAR-10, with five primary data configurations:
\begin{itemize}
    \item SDSS: Clients have the same distribution of the same size data.
    \item DDSS: Clients with different distributions of data of the same size.
    \item SDDS: Clients with the same distribution of data of different size.
    \item NFSS: Clients have data of the same size from the same distribution, but there is some noise in the client's features.
    \item NLSS: Clients have data of the same size from the same distribution, but there is some noise in the client's labeling.
\end{itemize}
% 本实验采用的联邦聚合策略为FedAvg (Federated Averaging)，实验所采用的数据集为cifar-10，主要涉及如下5中数据设置。
% SDSS：

\begin{table}[h!]
\centering % 确保表格居中
\caption{Experiments Setting}
\begin{tabular}{m{3cm} m{2cm} m{2cm}} % 所有列都使用 m{}，允许换行并上下居中
\hline
\textbf{Dataset} & \textbf{Data Type} & \textbf{Model} \\ \hline
Iris\cite{fisher1936use}                          & \hspace{0.3cm}Value  & MLP      \\ \hline
CIFAR-10\cite{krizhevsky2009learning}, MNIST\cite{lecun1998gradient},\hspace{1cm} FASHION-MNIST\cite{xiao2017fashion}   & \hspace{0.3cm}Image  & ResNet18 \\ \hline
Movie Review Dataset\cite{maas-EtAl:2011:ACL-HLT2011}            & \hspace{0.3cm}Text   & LSTM     \\ \hline
\end{tabular}
\label{tab:experiments1}
\end{table}

\subsection{Simple sample scale in different dataset}
% 为了验证各种类型的数据集上简单样例的规模比例，我们在数值、图像和文本类型的数据集上训练了相应的模型。在本实验中，我们采用单机训练的模式，并将训练集和测试集按 8:2 的比例划分。同时，简单样例的判定标准如下：在 30 个连续的训练周期中被判定为正确的样例被认定为简单样例。实验其余设置如\ref{}所示。
To validate the proportional scale of simple examples across various types of datasets, we trained corresponding models on numerical, image, and text datasets, recording the results of each validation round.  In this experiment, we employed a single-machine training mode and split the dataset into training and testing sets with an 8:2 ratio.  The criterion for identifying simple examples was as follows: samples correctly classified for 30 consecutive training epochs were designated as simple examples. The remaining experimental settings are detailed in table\ref{tab:experiments1}

% \begin{table}[h!]
% \centering
% \caption{Experiments Setting}
% \begin{tabular}{|p{4cm}|p{1.5cm}|p{1.5cm}|} % 使用 p{宽度} 来定义可以换行的列
% \hline
% \textbf{Dataset} & \textbf{Data Type} & \textbf{Model} \\ \hline
% Iris\cite{fisher1936use}                          & Value  & MLP      \\ \hline
% CIFAR-10\cite{krizhevsky2009learning}, MNIST\cite{lecun1998gradient}, FASHION-MNIST\cite{xiao2017fashion}   & Image  & ResNet18 \\ \hline
% Movie Review Dataset\cite{maas-EtAl:2011:ACL-HLT2011}            & Text   & LSTM     \\ \hline
% \end{tabular}
% \label{tab:experiments1} % 为表格指定一个标签，以便在正文中引用
% \end{table}

\begin{figure}[H]
    \centering
    \includegraphics[width=1\linewidth]{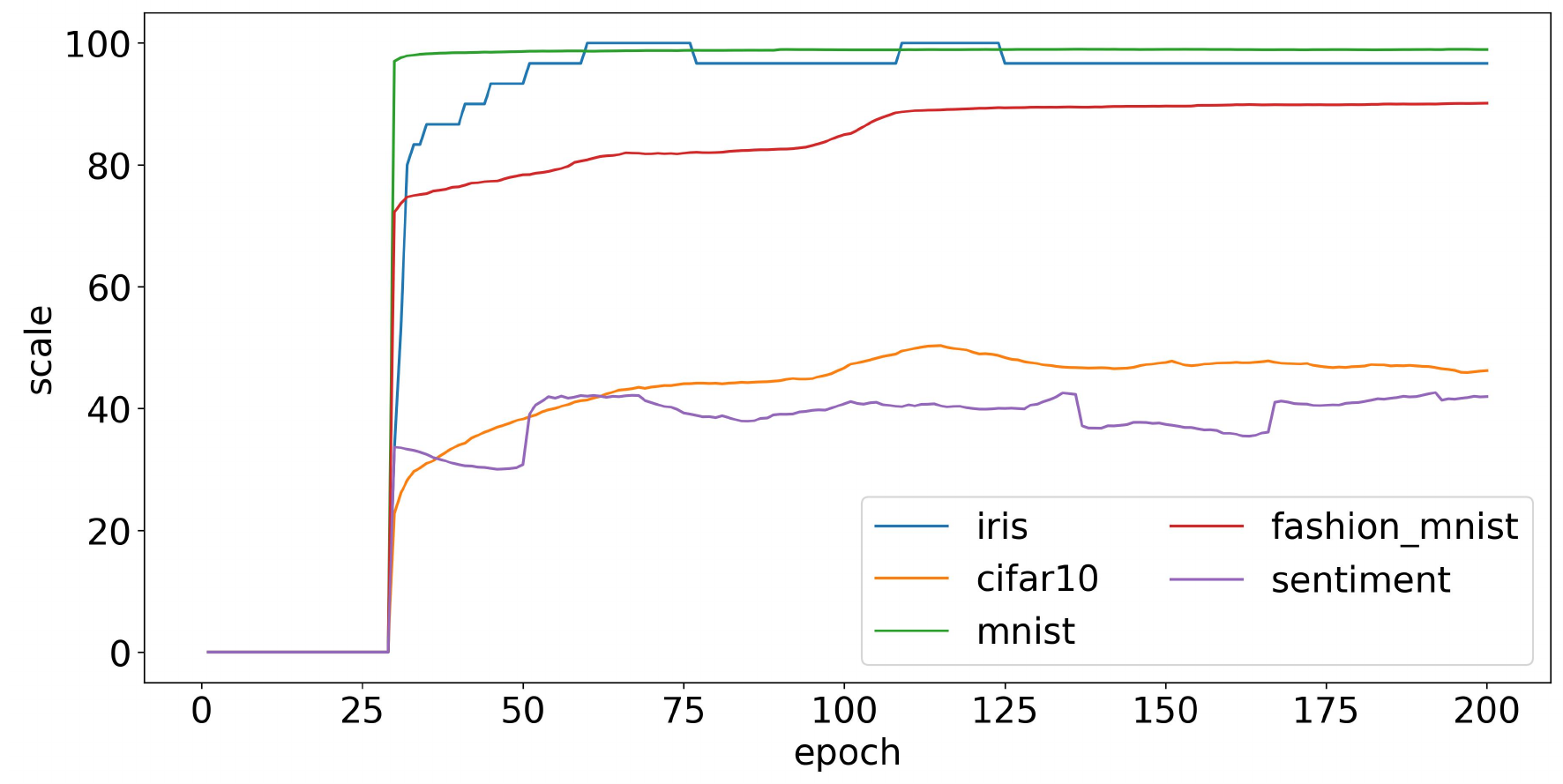}
    \caption{Simple sample scale in different dataset}
    \label{fig:experiment1}
\end{figure}

As shown in Figure \ref{fig:experiment1}, the experimental results demonstrate significant variations in the proportion of simple examples across different types and categories of datasets. Generally, this proportion ranges from 30\%+ to 90\%+, indicating that pruning the validation set is indeed meaningful and potentially beneficial.
% 实验结果如\ref{}所示，实验结果表明，对于不同类型和种类的数据集，不同数据集之间的比例可能存在较大差异。一般来说，这一比例范围从 30%+ 到 90%+，因此对验证集进行剪枝是有意义的。

\subsection{Time comparison of different strategies}
We have performed model training on the cifar-10 dataset after putting our strategy on the cifar-10 dataset with different strategies. The experimental results are shown in the following figure\ref{fig:experiment2}.The time per round of validation for our method in the early time is higher than the original full time, but as the accuracy of the model increases, the number of pruned samples increases, which offsets this consumption and saves more time. It can be seen that the overall picture is that the Ignore strategy drops the fastest in the time dimension. Since randomChoose will have a portion of regression, resulting in the consumption of time to process the random strategy and the testing of this portion of regression samples, randomChoose will be higher than the Ignore strategy. Since weightChoose has to perform weight randomization compared to randomChoose, this operation will consume more time compared to randomChoose.Concurrently, under the same number of rounds, the weightChoose pruning method eliminates fewer data points compared to randomChoose. This is because weightChoose demonstrates a superior ability to accurately identify samples that transition from being simple to challenging examples due to model modifications.

% 我们在同等条件下进行了该实验，实验的数据集为cifar-10数据集，模型的选择为ResNet50，唯一不同的是采取了两种不同的抽样策略，并将这两种抽样策略与全量验证进行对比。实验结果如下图所示。在早期阶段，我们random_choose和weight_choose策略的每轮验证时间会高于全量验证的时间，但随着模型准确率的提高，被剪枝样例的数量增加，剪枝样例减少的时间抵消了记录信息以及抽样带来的开销并节省了更多时间。总体来看，randomChoose策略在时间维度上下降速度快于weightChoose，这是由于weightChoose相比于randomChoose需要依据权重进行抽取，这一操作的时间消耗也会比randomChoose的抽取更加耗时。同时在相同轮次下，weightChoose剪枝的数据量少于randomChoose，因为weightChoose能够更精准的捕捉到因模型变动导致从简单变为困难样例的样例。

\begin{figure}[h]
    \centering
    \includegraphics[width=1.0\linewidth]{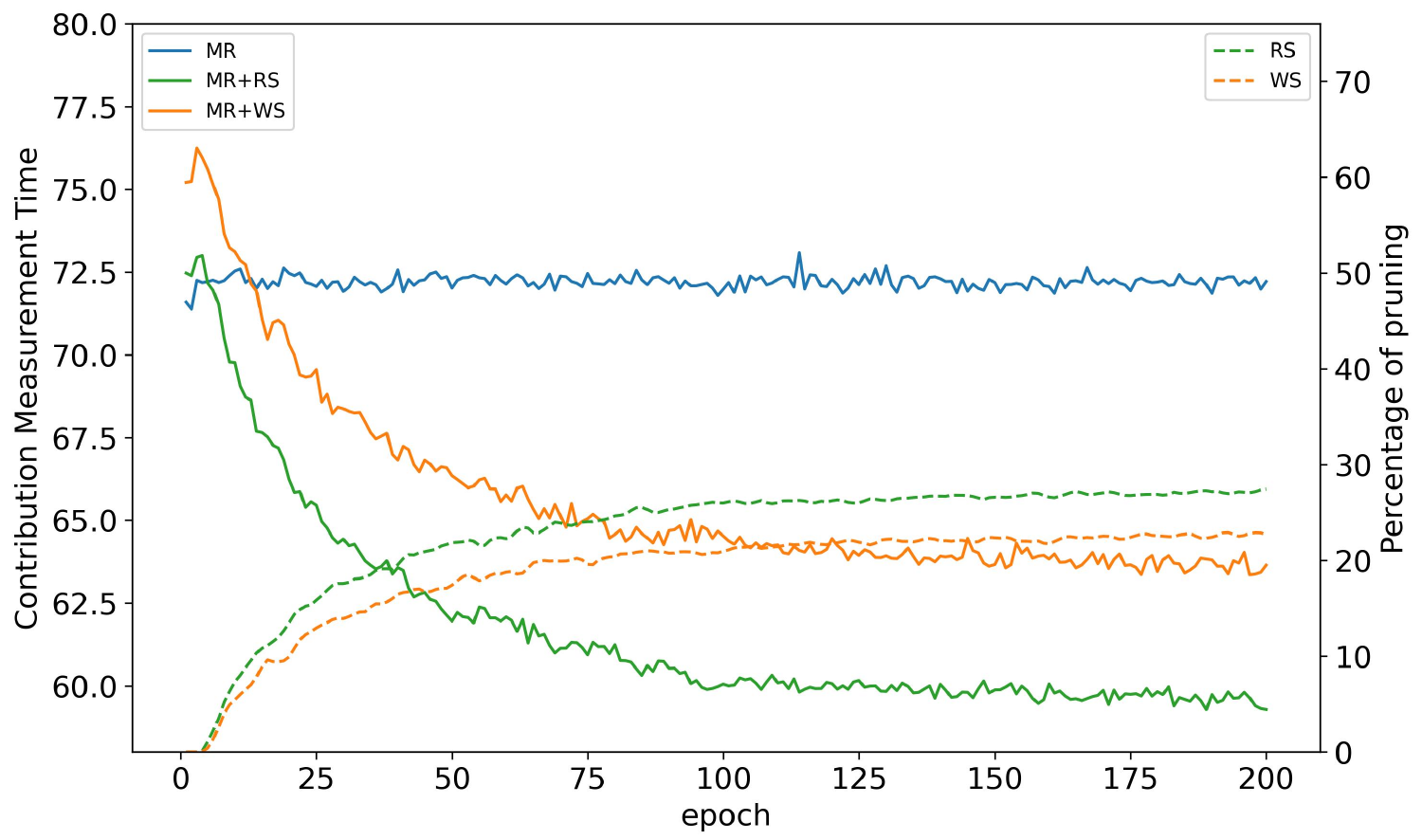}
    \caption{Evalution change under different strategies}
    \label{fig:experiment2}
\end{figure}

\subsection{Effect of aggregated gradient number on the accuracy of pruning data set}
To validate the accuracy of our method on federated learning, we average the cifar-10 data across 5 clients and validate it on the server. Since gradient combinations are needed to compute the sum of marginal effects across clients, we perform a difference measure between real and simulated accuracy for various length combinations. For this purpose, we set up controlled experiments to demonstrate the effectiveness of weighted sampling (WS) and the sampling proportion determined by the number of clients (CNDSR) in our method. For the RS and WS experiments we chose a regression proportion of 0.05 from the pruned sample. for CNDSR the aggregated number of clients 1,2,3,4,5, the regression proportions were taken to be 0.5,0.4,0.3,0.2,0.1 respectively.

% 为了验证我们的方法在联邦学习上的准确性，我们平均了 5 个客户端的 cifar-10 数据，在SDSS的分布下进行了验证。由于计算各客户端的边际效应之和需要梯度组合，因此我们对不同数量组合的实际准确度和模拟准确度进行了差异测量。为此，我们设置了对照实验，以证明加权采样（WS）和由客户端数量决定的采样比例（CNDSR）在我们的方法中的有效性。在 RS 和 WS 实验中，我们从剪枝样例中选择了 0.1 的回归比例。在 CNDSR 实验中，客户总数分别为 1、2、3、4、5，回归比例分别为 0.5、0.4、0.3、0.2、0.1。

\begin{figure}[h]
    \centering
    \includegraphics[width=1\linewidth]{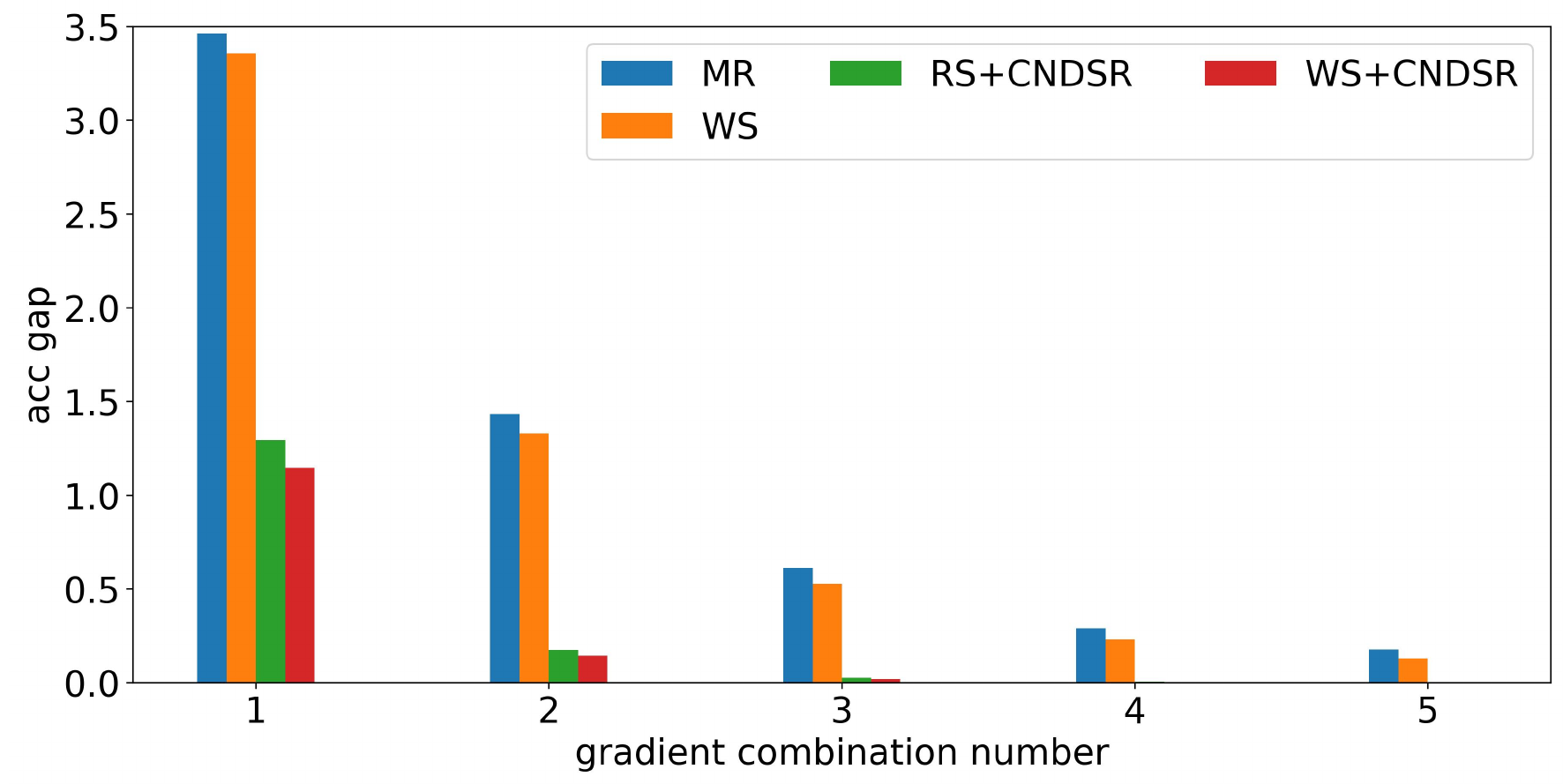}
    \caption{Gap between fitted accuracy and true accuracy}
    \label{fig:experiment3}
\end{figure}

The experimental results indicate that an insufficient number of combinations can cause the model parameters to excessively align with a single client's local data, leading to over-localization. This prevents the model from learning global features, resulting in significant model fluctuations and a substantial disparity between the fitted accuracy and the true accuracy, potentially impacting the contribution assessment results. When employing the CNDSR-based method, we observed that moderately increasing the proportion of samples helps reduce the accuracy gap. Similarly, the WS method also contributes to narrowing this discrepancy. With 3-5 clients, the CNDSR-based method achieves accuracy nearly identical to the true values. Consequently, when aggregating data from a small number of clients, we extract a larger proportion of samples, or even all samples, from the redundant dataset.

% 实验结果表明，当组合数过少时，会使得模型的参数过于趋向于某一客户端本地的数据，从而导致模型过于本地化，无法学习到全局特征，导致模型变化较大，拟合的准确率与真实准确率差距过大，可能会进一步影响模型贡献度评估结果。当我们采用基于CNDSR的方法时，我们发现适当扩大比例有利于准确率差距的减少，同时WS方法也能降低准确率的差距。当客户端数达到3-5时，基于CNDSR的方法在准确率上几乎与真实值一致。对此在聚合客户端数量较少时，我们会从冗余样例中提取较大比例的样例甚至全部样例。

\subsection{Tested on gradient aggregation-based MR}
Finally we applied our components on the Shapley value method based on gradient aggregation and simulated five data distribution conditions to verify the effectiveness of our method,and the five data settings are as above.
% 最后，我们在基于梯度聚合的 Shapley 值方法上应用了我们的组件，并模拟了五种数据分布条件，以验证我们方法的有效性，五种数据设置如上。

We adopt the gradient aggregation-based method MR, which is the earliest approach to calculate Shapley values for measuring participant contributions through gradient aggregation. Many subsequent methods have built upon and improved this approach. The main concept of this method is as follows: In each round, the server collects gradient updates from clients. By combining these gradients, relevant sub-models are reconstructed, and performance metrics are evaluated on these sub-models. Finally, the contribution of each party is determined by summing the marginal benefits across different combinations.
% 我们采用的基于梯度聚合的方法为MR，它是最早的基于梯度聚合来计算夏普利值从而衡量各方贡献的方法，后续有很多方法在他的基础上进行了改进。该方法的主要思想如下：在每一轮中，服务器会收集客户端更新的梯度。通过这些梯度的组合重建相关的子模型，并在子模型进行指标衡量。最终根据组合间的边际效益之和来获得各方的贡献度。

Evaluation Metrics.We compare different methods on the following metrics.
\begin{itemize}
    \item Time Saving: Calculate the percentage of time saved by each party's contribution.
    \item Cosine Distance: Let the vectors of normalized contribution index of different data providers calculated according to the True Shapley and by an approximated method is denoted by $\phi^* = \langle \phi_1^*, \phi_2^*, \cdots, \phi_n^* \rangle \quad \text{and} \quad \phi = \langle \phi_1, \phi_2, \cdots, \phi_n \rangle$, respectively. The Cosine Distance is defined by
    $$d_{\text{cos}}(\phi^*, \phi) = 1 - \frac{\sum_{i=1}^{n} \phi_i^* \cdot \phi_i}{\sqrt{\sum_{i=1}^{n} (\phi_i^*)^2} \sqrt{\sum_{i=1}^{n} (\phi_i)^2}}$$
    \item Euclidean Distance: The Euclidean Distance is defined by
    $$d_{\text{euclid}}(\phi^*, \phi) = \sqrt{\sum_{i=1}^{n} (\phi_i^* - \phi_i)^2}$$
    \item Maximum Difference: The Maximum Difference is defined by
    $$d_{\text{max}}(\phi^*, \phi) = \max_{1 \leq i \leq n} |\phi_i^* - \phi_i|$$
\end{itemize}

% 实验的其他设置如下：在EE中针对不同数量的组合客户端数的抽样比例为：[1.0,1.0,0.5,0.1,0.1],在[]
The other settings of the experiment were as follows:
\begin{itemize}
    \item RD:Random sample from easy set.
    \item WR:Random selection by weight from easy set.
    \item \lbrack1.0,1.0,0.5,0.1,0.1\rbrack is the extraction ratios for different length combinations in EE.
    \item \lbrack0.1,0.1,0.1,0.1,0.1\rbrack is the extraction ratios for different length combinations in ET.
\end{itemize}

\begin{figure}[h]
    \centering
    \includegraphics[width=1.0\linewidth]{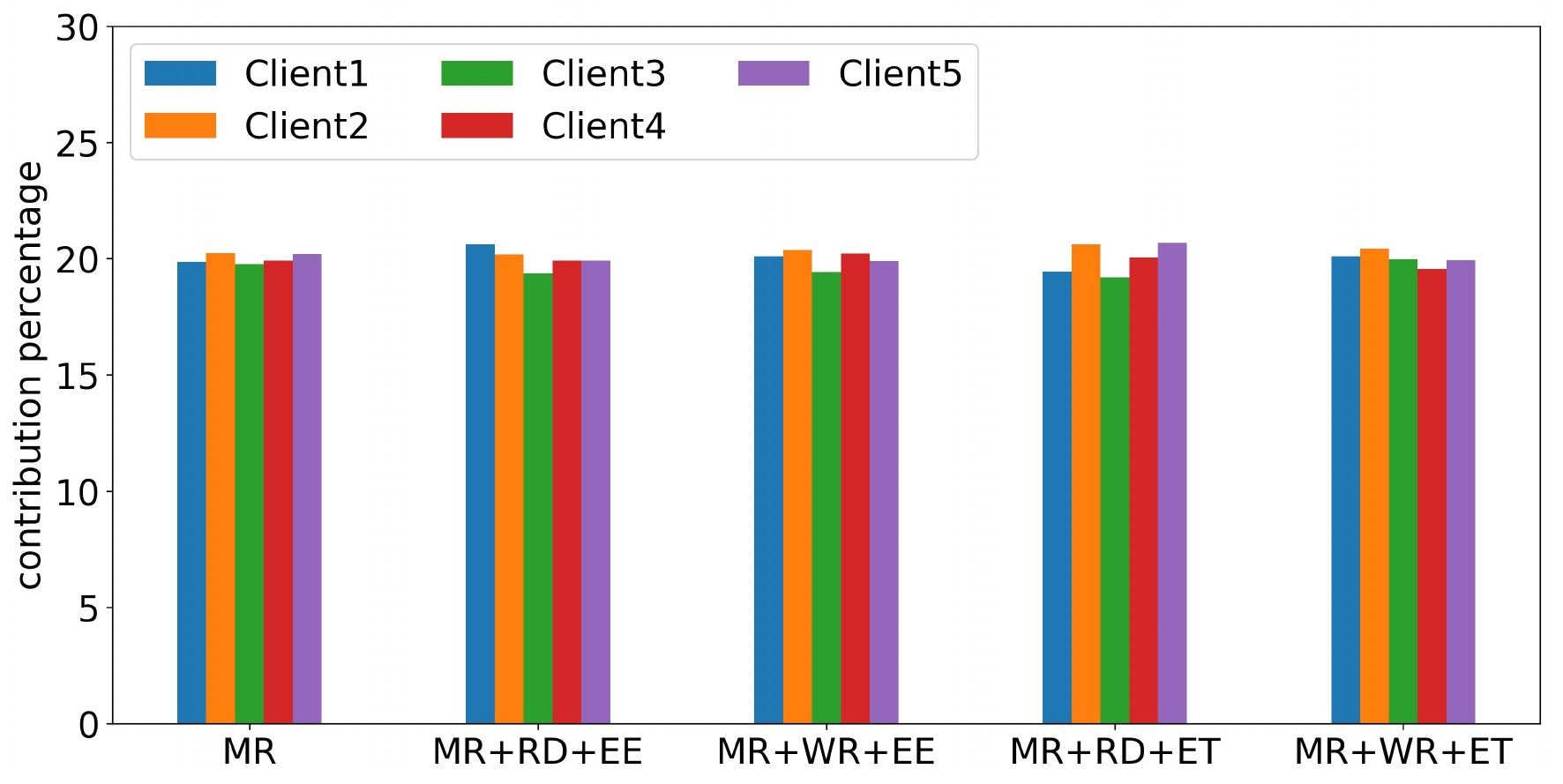}
    \caption{Contributions under sdss}
    \label{fig:contributions_sdss}
\end{figure}

\begin{figure}[h]
    \centering
    \includegraphics[width=1.0\linewidth]{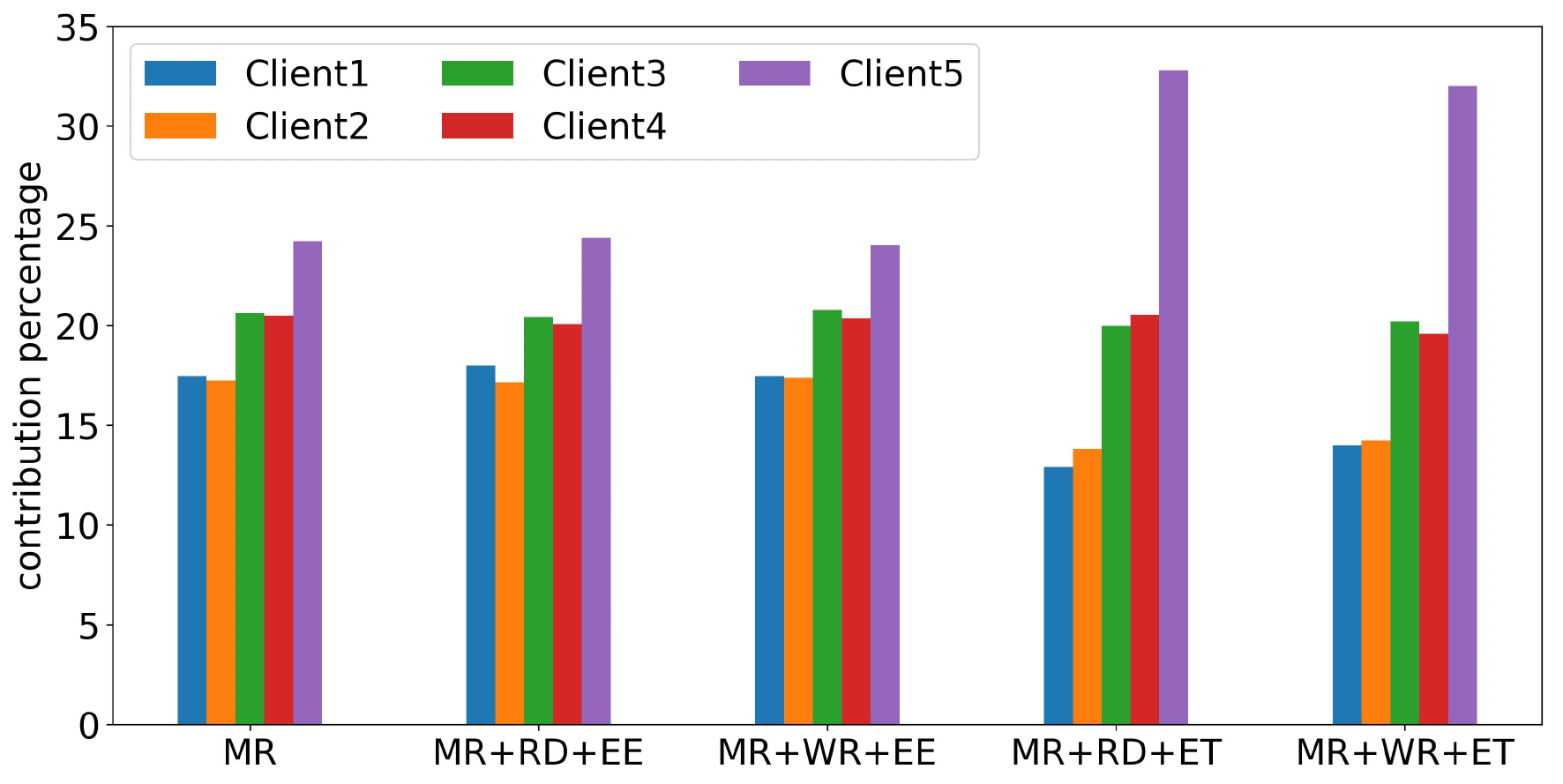}
    \caption{Contributions under sdds}
    \label{fig:contributions_sdds}
\end{figure}

\begin{figure}[h]
    \centering
    \includegraphics[width=1.0\linewidth]{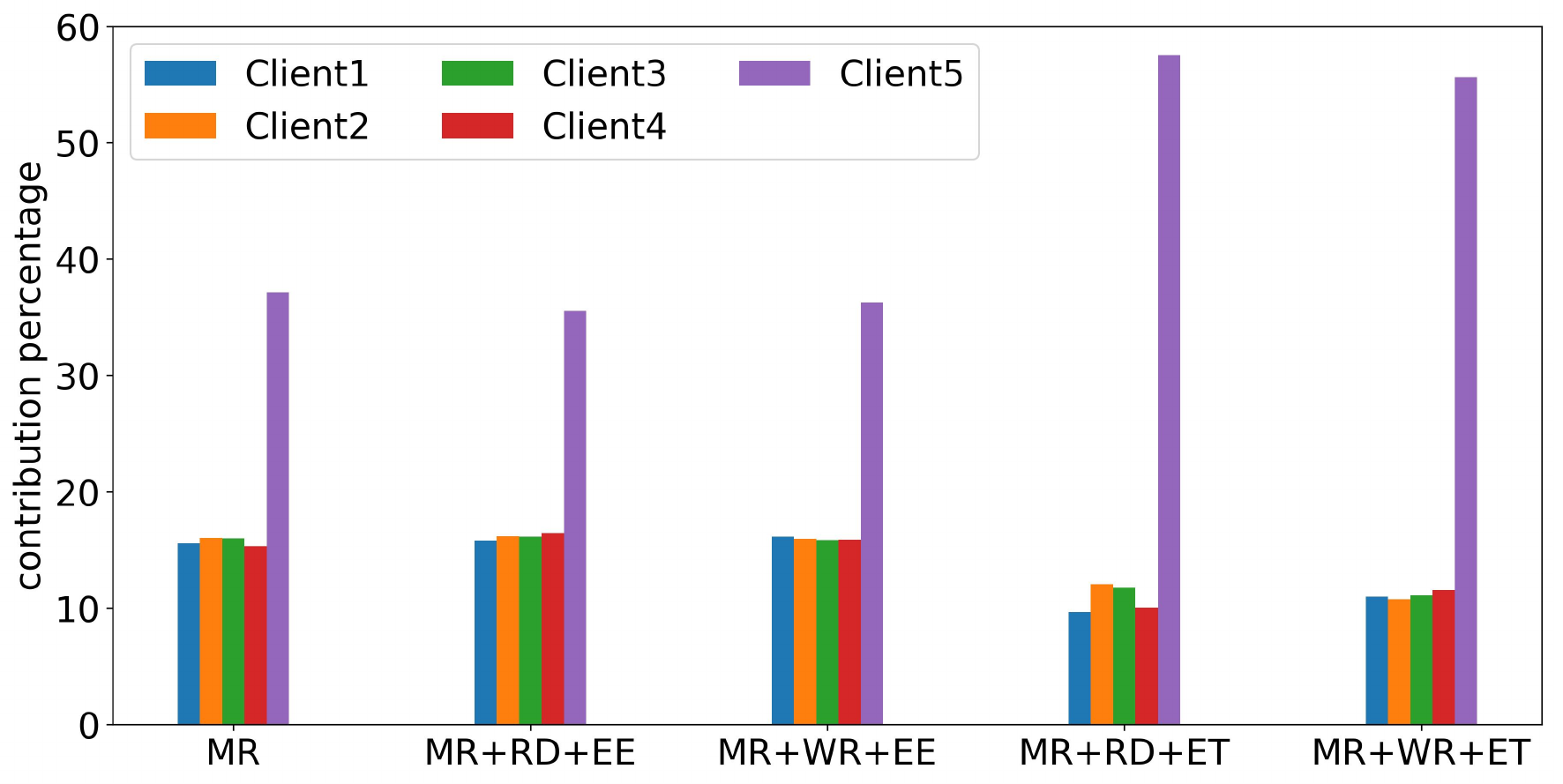}
    \caption{Contributions under ddss}
    \label{fig:contributions_ddss}
\end{figure}

\begin{figure}[h]
    \centering
    \includegraphics[width=1.0\linewidth]{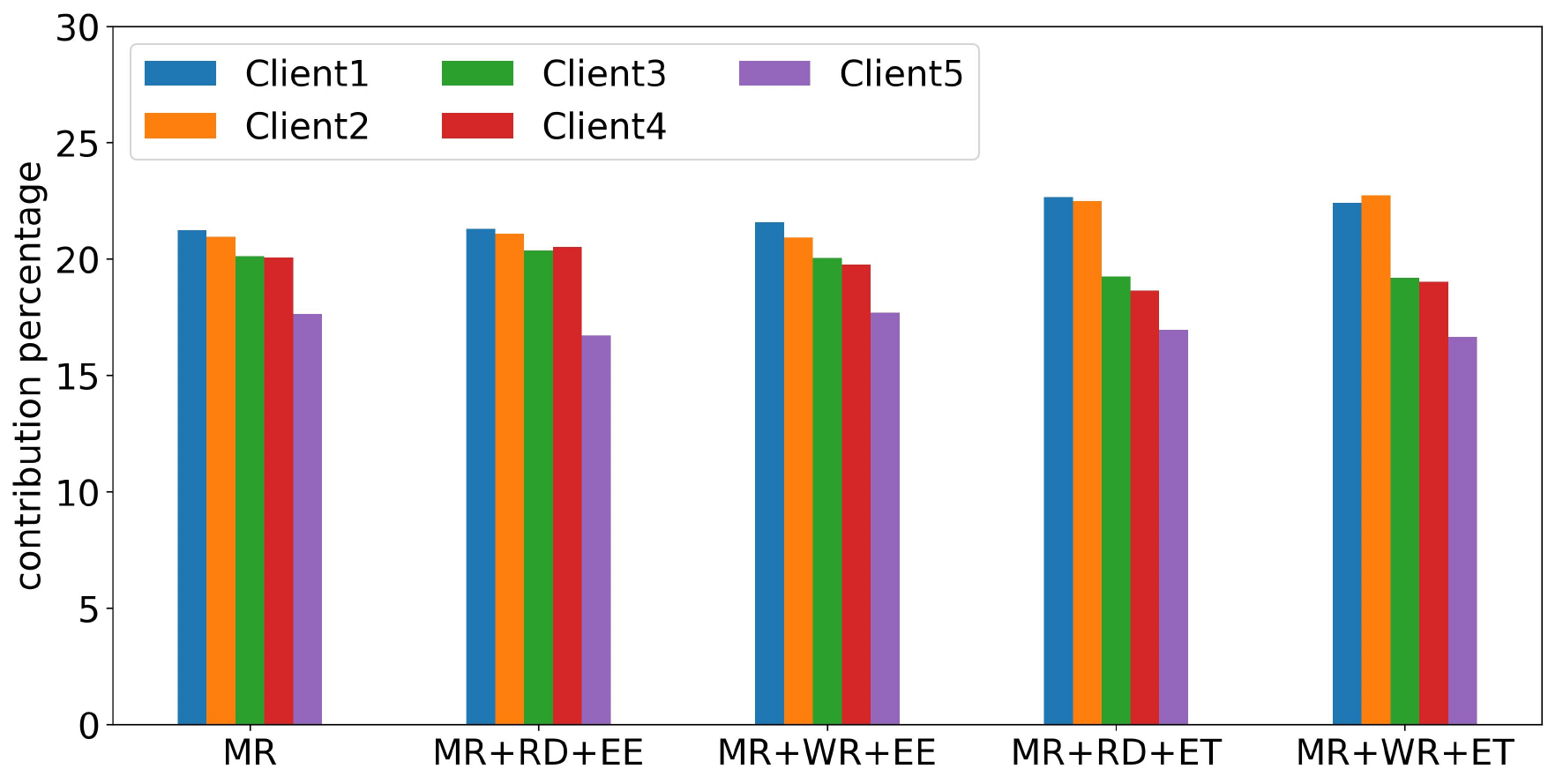}
    \caption{Contributions under nfss}
    \label{fig:contributions_nfss}
\end{figure}

\begin{figure}[h]
    \centering
    \includegraphics[width=1.0\linewidth]{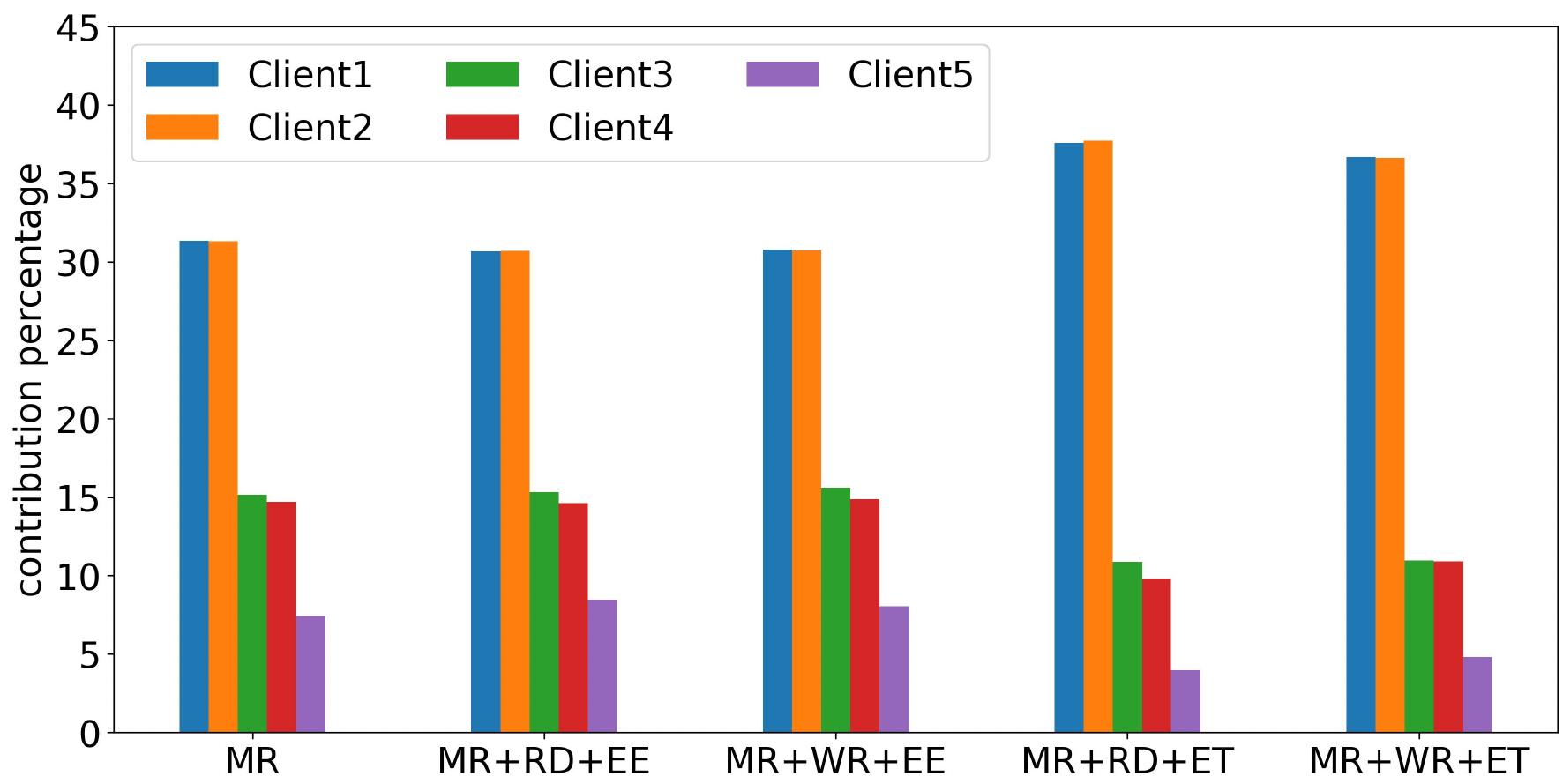}
    \caption{Contributions under nlss}
    \label{fig:contributions_nlss}
\end{figure}

\begin{table}[h]
\centering
\caption{Metrics under SDSS}
\label{tab:metrics_under_sdss} % 为表格指定一个标签，以便在正文中引用

\begin{tabular}{>{\centering\arraybackslash}m{1.5cm}>{\centering\arraybackslash}m{1.0cm}>{\centering\arraybackslash}m{1.3cm}>{\centering\arraybackslash}m{1.3cm}>{\centering\arraybackslash}m{1.5cm}}\hline
\textbf{} & \textbf{Time Saving} & \textbf{Cosine Distance} & \textbf{Euclidean Distance} & \textbf{Maximum Difference} \\ \hline
MR+RD+EE &    10.47591     &  0.000210  & 0.916187  & 0.760000      \\ \hline
MR+WR+EE &     9.78873     &  0.000102  & 0.637495  & 0.359999      \\ \hline
MR+RD+ET &    34.09605     &  0.000219  & 0.937389  & 0.570000      \\ \hline
MR+WR+ET &    32.20264     &  0.000079  & 0.565066  & 0.339999      \\ \hline
\end{tabular}
\end{table}

\begin{table}[h]
\centering
\caption{Metrics under DDSS}
\label{tab:metrics_under_ddss} % 为表格指定一个标签，以便在正文中引用
\begin{tabular}{>{\centering\arraybackslash}m{1.5cm}>{\centering\arraybackslash}m{1.0cm}>{\centering\arraybackslash}m{1.3cm}>{\centering\arraybackslash}m{1.3cm}>{\centering\arraybackslash}m{1.5cm}}\hline
\textbf{} & \textbf{Time Saving} & \textbf{Cosine Distance} & \textbf{Euclidean Distance} & \textbf{Maximum Difference} \\ \hline
MR+RD+EE &    12.40040    &  0.000750  & 1.995294   & 1.630000      \\ \hline
MR+WR+EE &    7.51847     &  0.000270  & 1.192308   & 0.890000      \\ \hline
MR+RD+ET &    28.58902    &  0.057952  & 22.649048  & 20.40000      \\ \hline
MR+WR+ET &    26.78545    &  0.051820  & 20.689838  & 18.47999      \\ \hline
\end{tabular}
\end{table}

\begin{table}[h]
\centering
\caption{Metrics under SDDS}
\label{tab:metrics_under_sdds} % 为表格指定一个标签，以便在正文中引用

\begin{tabular}{>{\centering\arraybackslash}m{1.5cm}>{\centering\arraybackslash}m{1.0cm}>{\centering\arraybackslash}m{1.3cm}>{\centering\arraybackslash}m{1.3cm}>{\centering\arraybackslash}m{1.5cm}}\hline
\textbf{} & \textbf{Time Saving} & \textbf{Cosine Distance} & \textbf{Euclidean Distance} & \textbf{Maximum Difference} \\ \hline
MR+RD+EE &   12.51104     &  0.000121  & 0.702139  & 0.500000      \\ \hline
MR+WR+EE &   11.84869     &  0.000026  & 0.330151  & 0.199999      \\ \hline
MR+RD+ET &   31.24816     &  0.023610  & 10.329632 & 8.579999      \\ \hline
MR+WR+ET &   27.17839     &  0.018510  & 9.076012  & 7.770000      \\ \hline
\end{tabular}
\end{table}

\begin{table}[h!]
\centering
\caption{Metrics under NFSS}
\label{tab:metrics_under_nfss} % 为表格指定一个标签，以便在正文中引用

\begin{tabular}{>{\centering\arraybackslash}m{1.5cm}>{\centering\arraybackslash}m{1.0cm}>{\centering\arraybackslash}m{1.3cm}>{\centering\arraybackslash}m{1.3cm}>{\centering\arraybackslash}m{1.5cm}}\hline
\textbf{} & \textbf{Time Saving} & \textbf{Cosine Distance} & \textbf{Euclidean Distance} & \textbf{Maximum Difference} \\ \hline
MR+RD+EE &   12.68942     &  0.000278  & 1.060188  & 0.910000      \\ \hline
MR+WR+EE &    7.24584     &  0.000055  & 0.472652  & 0.349999      \\ \hline
MR+RD+ET &   33.20582     &  0.001881  & 2.760706  & 1.529999      \\ \hline
MR+WR+ET &   25.24643     &  0.001841  & 2.733148  & 1.779999      \\ \hline
\end{tabular}
\end{table}

\begin{table}[h!]
\centering
\caption{Metrics under NLSS}
\label{tab:metrics_under_nlss} % 为表格指定一个标签，以便在正文中引用

\begin{tabular}{>{\centering\arraybackslash}m{1.5cm}>{\centering\arraybackslash}m{1.0cm}>{\centering\arraybackslash}m{1.3cm}>{\centering\arraybackslash}m{1.3cm}>{\centering\arraybackslash}m{1.5cm}}\hline
\textbf{} & \textbf{Time Saving} & \textbf{Cosine Distance} & \textbf{Euclidean Distance} & \textbf{Maximum Difference} \\ \hline
MR+RD+EE &   10.61531     &  0.000314  & 1.387372  & 1.010000      \\ \hline
MR+WR+EE &    6.10894     &  0.000219  & 1.137365  & 0.600000      \\ \hline
MR+RD+ET &   30.14041     &  0.018381  &11.567329  & 6.399999      \\ \hline
MR+WR+ET &   24.26670     &  0.013626  & 9.746902  & 5.309999      \\ \hline
\end{tabular}
\end{table}
In the EE experiments, we aimed to closely match the performance of the MR method. As evidenced by the data in Tables 2-6 and Figures 5-9, our method achieves nearly identical fitting results compared to the original MR method. This observation is further corroborated by various metrics including Cosine Distance, Euclidean Distance, and Maximum Difference. Notably, we achieved a 9\% to 14\% improvement in speed, as indicated by the Time Saving metric.The success of our method in accurately simulating the original dataset's accuracy can be attributed to two key strategies: timely updates and setting higher sampling ratios for fewer combinations. These findings not only validate our experimental hypothesis but also demonstrate that the MR method can achieve speed improvements within an acceptable margin of error.
%在EE实验中,我们致力于尽可能地匹配MR方法的效果。通过表格2-6和图表5-9的数据,我们可以看出我们的方法在拟合效果上几乎与原有MR方法达到一致。这一观点同时也得到了Cosine Distance、Euclidean Distance和Maximum Difference等衡量指标的侧面佐证。值得注意的是,在Time Saving指标上,我们实现了6%~12%的速度提升。我们的方法之所以能在模拟原有数据集的准确率上取得较好效果,主要归功于两个策略:及时更新和对较少组合设置较高的抽样比例。这不仅印证了我们的实验假设,同时也使MR方法在可容忍的误差范围内实现了速度的提升。

When shifting our focus from replicating the MR method to assigning higher weights to difficult samples, the ET strategy becomes a preferable choice. The ET strategy amplifies the disparities in contributions among parties, enabling a more distinct tiered ranking of contributions for stakeholders primarily concerned with contribution rankings. Consider the nfss data configuration illustrated in Figure 8: Clients 1 and 2 have unaltered data features, Clients 3 and 4 have 15\% Gaussian noise added to their data, and Client 5 has 30\% noise added. In this scenario, the conventional MR method struggles to clearly differentiate between Clients 1, 2 and Clients 3, 4 due to their closely aligned contribution values. This similarity makes it challenging to definitively assert that the data quality of Clients 1 and 2 is superior to that of Clients 3 and 4, especially considering the inherent randomness in model training processes.
By employing the ET strategy, we set lower regression ratios across different combinations, resulting in a pruned dataset with a higher proportion of difficult samples, thereby increasing the weight of these challenging examples. This approach successfully widens the gap in contribution percentages among parties with varying data quality during the testing phase, leading to a more effective contribution ranking. Moreover, this method significantly enhances the speed of contribution assessment, achieving an acceleration effect of over 30\%+.
%当我们不再专注于拟合MR方法,而是希望赋予难例样本更高权重时,可以选用ET策略。ET策略拉大了各方贡献度之间的差距,使得更关注贡献度排名的需求方可以更明确地进行贡献度的阶梯化排名。以图8所示的nfss数据设置为例:客户端1、2的数据特征没有添加噪声,客户端3、4的数据添加了15%的高斯噪声,客户端5则添加了30%的噪声。在这种情况下,常规的MR方法难以明显区分客户端1、2与客户端3、4,因为它们的贡献值十分接近。这种接近使得我们无法断定客户端1、2的数据质量一定优于客户端3、4,尤其考虑到模型训练过程中存在一定的随机性。而当我们使用ET策略时,模型在不同组合中均设置较低的回归比例,这使得剪枝后的数据集中困难样本比例较高,从而加重了困难样本的权重。这一做法在测试过程中成功拉开了拥有不同质量数据的各方的贡献度占比,从而实现了更好的贡献度排名效果。此外,这种方法还显著提高了贡献度评估的速度,达到了30%以上的加速效果。

\section{Conclusion}
This paper addresses the challenge of enhancing contribution assessment efficiency in federated learning (FL) systems. We introduce the Dynamic Validation Pruning Set Shapley (DPVS-Shapley) method, which employs a dynamic pruning strategy to categorize the validation set into simple and difficult samples. By selectively regressing different proportions of simple samples, we achieve two objectives: reducing overall validation time without compromising accuracy, and amplifying the contribution disparities among parties by assigning higher weights to difficult samples while significantly shortening validation time.

\begin{acks}
If you wish to include any acknowledgments in your paper (e.g., to 
people or funding agencies), please do so using the `\texttt{acks}' 
environment. Note that the text of your acknowledgments will be omitted
if you compile your document with the `\texttt{anonymous}' option.
\end{acks}

%%%%%%%%%%%%%%%%%%%%%%%%%%%%%%%%%%%%%%%%%%%%%%%%%%%%%%%%%%%%%%%%%%%%%%%%

%%% The next two lines define, first, the bibliography style to be 
%%% applied, and, second, the bibliography file to be used.

\bibliographystyle{ACM-Reference-Format} 
\bibliography{sample}

%%%%%%%%%%%%%%%%%%%%%%%%%%%%%%%%%%%%%%%%%%%%%%%%%%%%%%%%%%%%%%%%%%%%%%%%

\end{document}